\documentclass{article}

\PassOptionsToPackage{numbers}{natbib}
\usepackage[preprint]{neurips_2025}

\usepackage[utf8]{inputenc} 
\usepackage[T1]{fontenc}    
\usepackage{hyperref}       
\usepackage{url}            
\usepackage{booktabs}       
\usepackage{amsfonts}       
\usepackage{nicefrac}       
\usepackage{microtype}      
\usepackage{xcolor}         

\usepackage{amssymb}
\usepackage{multirow}
\usepackage{pgfplots}
\usepackage{float}
\usepackage{filecontents}
\usepackage{graphicx}
\usepackage{placeins}
\usepackage{eurosym}
\usepackage{pgfplotstable}
\usepackage{breakcites}
\usepackage{amsmath}
\usepackage{cleveref}
\usepackage{mathtools}
\usepackage{physics}
\usepackage{tikz}     
\usepackage{graphicx}
\usepackage{color}
\usepackage{microtype}
\usepackage{comment}
\usepackage{tcolorbox}
\usepackage{ragged2e}
\usepackage{soul}

\colorlet{lightblue}{blue!15}
\colorlet{lightgreen}{green!15}
\colorlet{lightred}{red!15}
\colorlet{lightyellow}{yellow!10!white}

\usetikzlibrary{positioning}
\usetikzlibrary{calc}

\usetikzlibrary{external}
\title{Exploring the generalization of LLM truth directions on conversational formats}

\author{%
  Timour Ichmoukhamedov \quad David Martens \\
  ADM, Universiteit Antwerpen \\
  Prinsstraat 13, 2000 Antwerp, Belgium \\
  \texttt{timour.ichmoukhamedov@uantwerpen.be} 
}

\begin{document}

\maketitle

\begin{abstract}
    Several recent works argue that LLMs have a universal \textit{truth direction} where true and false statements are linearly separable in the activation space of the model. It has been demonstrated that linear probes trained on a single hidden state of the model already generalize across a range of topics and might even be used for lie detection in LLM conversations. In this work we explore how this truth direction generalizes between various conversational formats. We find good generalization between short conversations that end on a lie, but poor generalization to longer formats where the lie appears earlier in the input prompt. We propose a solution that significantly improves this type of generalization by adding a fixed key phrase at the end of each conversation. Our results highlight the challenges towards reliable LLM lie detectors that generalize to new settings.
\end{abstract}

\section{Introduction}\label{sec:introduction}
 
\begin{figure*}[!htbp]
\centering
\makebox[\textwidth][c]{%
\resizebox{\textwidth}{!}{%
\includegraphics[width=\textwidth]{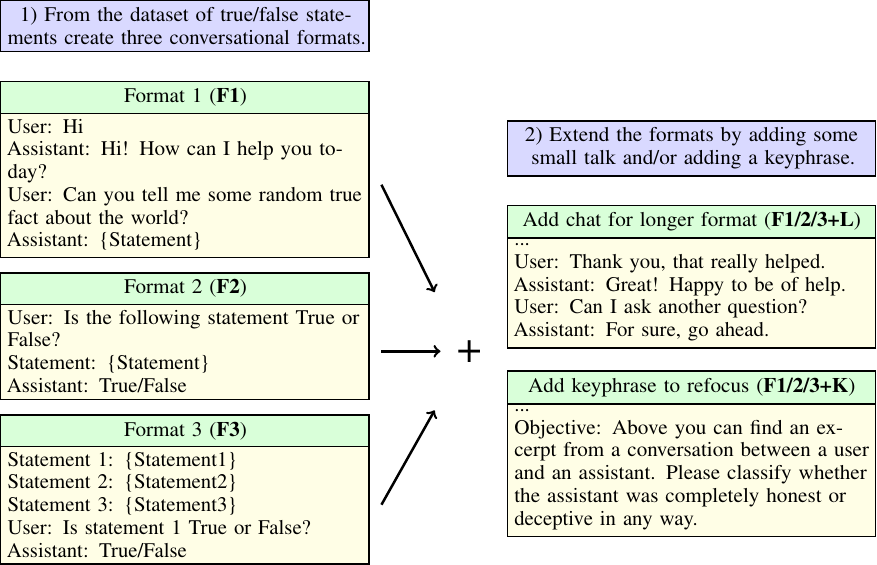}
}
}
\caption{A visual overview of the various conversation formats used in this work. In total, we create 12 conversational datasets by adding three types of extensions on top of the three base conversational formats: 3$\times$(F, F+L, F+K, F+L+K).}
\label{fig:Figure1}
\end{figure*}

In light of the increasing concerns about AI deception~\cite{PARK2024100988,Hagendorff,scheurer2024large}, it is crucial to understand how LLMs internally represent and process concepts such as truth and lies. With the rapid and remarkable advances in the field of mechanistic interpretability~\cite{lindsey2025biology}, this naturally raises the question if one can create LLM lie detectors that are reliable in new and complex scenarios related to AI safety.

The past couple of years have indeed seen significant progress towards this goal. First, it has been demonstrated that various types of probes can be trained on the internal activations of LLMs to detect truths and falsehoods \cite{burns2023discovering,azaria}. This indicates that truth and lies are naturally emergent classes that LLMs internally represent. Around the same time, similar ideas have also been explored to detect LLM hallucinations \cite{ch-wang-etal-2024-androids, ji-etal-2024-llm}.
These approaches have since then also been generalized to detect a much broader class of LLM behaviors, including the possibility to engineer and steer those behaviors by nudging the activation space in a given direction \cite{zou2025representationengineeringtopdownapproach, li2023inferencetime}. From a more theoretical perspective, it has also been shown that lies in LLMs have a more complex signature from an information point of view and can be detected in entropy based metrics \cite{dombrowski2024an}. Another promising direction to detect lies in LLMs does not require access to the internal states, but shows that certain responses to follow-up questions are strongly correlated with a lie being told earlier \cite{pacchiardi2023catchailiarlie}.

Before proceeding, it is important to briefly address the loadedness of the term \textit{lie}, which implies a certain intentionality. For example, in one of the aforementioned works \cite{pacchiardi2023catchailiarlie}, lie was defined as outputting false statements despite knowing the truth. However, other works that will be discussed shortly, either avoid the term altogether or simply use it when discussing falsehoods. The use of this term is further obfuscated in the context of our work since  we will be studying the activations of fictitious LLM responses prepared in advance, that have not actually been generated by an LLM. Therefore, we want to clarify that in the rest of this work, the term \textit{lie} is for simplicity used synonymously with falsehood, and we do not ascribe to it any sense of intentionality.  From a practical lie detection perspective this approximation implies that the detector would catch all conversations from the LLM that are internally categorized as containing falsehoods, regardless of the intentionality. Whether LLMs produce intentional falsehoods in this context is discussed in greater detail in \cite{liu-etal-2023-cognitive}. 

The concept of a truth direction has been first introduced in \cite{marks2024the}, where the authors demonstrate the emergence of a clear linear separation between true and false input statements within a single hidden state of an LLM. A more recent work explores a series of further questions in this direction, including the question of generalization \cite{burger2024truth}. In particular, the authors present evidence that quite good generalization can be observed from probes trained on true/false statements to conversational prompts where the LLM is lying. However, this has been tested on a smaller sample of structurally similar conversations that all had a truth or a lie at the end.

In this work, we aim to explore how the truth direction introduced in \cite{marks2024the} generalizes between different types conversational formats, especially those where the lie appears earlier in the prompt. We present the following three main results.

\begin{itemize}
    \item \textbf{More evidence for generalization:} Linear probes trained on true/false statements generalize quite well to short conversational formats where the LLM tells a lie at the end. This supports the observations made in \cite{burger2024truth}.
    \item \textbf{Failures of generalization:} Linear probes trained either on statements or short conversations generalize very poorly to longer conversations where the lie does not occur at the very end. This challenges the previous observation and suggests that broad generalization of the truth direction across formats might be more challenging.
    \item \textbf{Improving generalization:} We demonstrate a significant improvement of this generalization by adding a fixed closing statement (key phrase) to each prompt that asks the model to think about truths and lies in the conversation.
\end{itemize}

It is important to emphasize that improving accuracies or performance compared to other linear probes is not the aim of this work. Criticism related to the generalization of the initial probes introduced in \cite{azaria} has already been raised in the past \cite{Levinstein2024}. However, this was related to generalizations from affirmative statements to negated statements, which has then subsequently been addressed in \cite{marks2024the}. With our work we hope to achieve a similar goal and point out potential pitfalls and remaining challenges to be tackled in this domain. 

\section{Methodology}\label{sec:Methodology}

The technical implementation of our approach is in many aspects similar to \cite{marks2024the,burger2024truth} and we will keep this part concise. For all further details we refer to our paper repository \footnote{\url{https://github.com/ADMAntwerp/llm-liedetector-keyphrase}}.

\textbf{Models:} We use two popular open-source LLMs -- \textit{LLama-3-8b-Instruct} \cite{llama3modelcard}, and \textit{Ministral-8b-Instruct-2410} \cite{mistral2410}. Confirming our observations on two independent models is sufficient for our aims, since linear separation of truths and lies has already been explored across several other models in \cite{marks2024the,burger2024truth}. The particular choice to include \textit{LLama-3-8b-Instruct} is because this model has been used in \cite{burger2024truth} to make the initial observation about generalization from statements to other prompts.  

\textbf{Creating input data:} We start from the datasets of single sentences labeled as true/false (e.g.~Paris is in Italy, label: false) obtained from \cite{burger2024truth} that also include data from previous works \cite{azaria,marks2024the,Levinstein2024}. The datasets cover 6 topics (e.g.~cities, facts, ...) and also include both affirmative and negated statements. From this dataset of statements we create additional conversational datasets in different formats where an assistant either tells a truth or a lie as depicted in Fig.~\ref{fig:Figure1}. For brevity, we introduce several abbreviations that we summarize in Table~\ref{table:table1}. It is important to emphasize that the main difference in the longer conversations is that they do not contain an LLM lie right at the very end. In what follows we will be mainly looking at the generalization from a short format to a longer format (F $\rightarrow$ F+L), and we will show that adding a key phrase at the end of the conversation (F+K $\rightarrow$ F+L+K) yields major improvements.

\begin{table*}[h!]
    \centering
    \begin{tabular}{c p{8cm}}  
    \toprule
    Statements & True/false statements from \cite{burger2024truth}. \\
    \midrule
    F1/F2/F3 & Base conversations with a lie at the end. \\
    \midrule
    F+L & Longer conversations with a lie earlier in prompt. \\
    \midrule
    F+K & Base conversations with a key phrase added. \\
    \midrule
    F+L+K & Longer conversations with a key phrase added. \\
    \bottomrule
    \end{tabular}
    \caption{Reference table for the various dataset formats.}
\end{table*}\label{table:table1}


\textbf{Collecting activations:} To collect the LLM activations for the input data created in the previous step, we follow a similar approach to \cite{burger2024truth}. For a given input, the last hidden state within a specific layer is recorded, which yields a dataset of 4096-dimensional vectors for every conversation instance. In \cite{burger2024truth} this was done for the hidden states of layer 12. However, since we will be dealing with longer inputs we also explore several layers further up to layer 20. We collect the activations for every instance in the previously discussed datasets (statements/formats/longer formats ...) and proceed to the next step. Collecting all the activations took around 4 hours per LLM (Llama and Mistral) on an A100 on Google Colab. 

\textbf{Centering the activations:} We are mainly interested in similar questions to \cite{marks2024the}, related to a universal truth \textit{direction} or \textit{plane}, and less interested in bias shifts of this plane which are likely to be prompt-dependent. For this reason, we also follow a similar approach to \cite{marks2024the}, and for every format, and every topic within that format, center all activations at the origin. This allows us to isolate the orientation of a potentially universal truth plane and explore its generalization. 

\textbf{Evaluation:} To evaluate the generalization of the truth direction, we train a simple linear regression probe on these centered datasets (LRC). For the evaluation we compute a cross-validation accuracy across various topics present in the datasets, just like in \cite{burger2024truth}. For example, consider that we want to compute how well the truth direction learned from dataset F2 generalizes to F3. We proceed to select a test set from F3 only containing statements on the topic of \textit{cities} and train a model on the training set in F2 where this topic has been removed. This is done across all topics and the average accuracy is then reported. During this procedure, the training set is also being actively balanced to contain an equal number of data points per topic. This part introduces some randomness and hence we run this cross-validation 10 times for every experiment.

\textbf{Overfitting concerns}: It is important to address a potential concern due to the relatively small dataset sizes used in previous (and our) works. In total, we have 5202 true/false statements which sets the dataset size of every other format. The dataset size is further reduced during evaluation when performing cross validation and active balancing. The training sets are therefore smaller than the dimensionality of the hidden state (4096) and one might have a natural concern whether our conclusions related to generalizations are sufficiently robust. To make sure that our conclusions hold up, we have rerun some experiments where we only trained on the first 100 PCA features of the training set. While the generalization values slightly improved, our three main conclusions remain fully valid.

\section{Results}\label{sec:Results}

\begin{figure*}[htbp!]
    \centering
    \makebox[\textwidth][c]{%
    \resizebox{\textwidth}{!}{%
    \includegraphics[width=\textwidth]{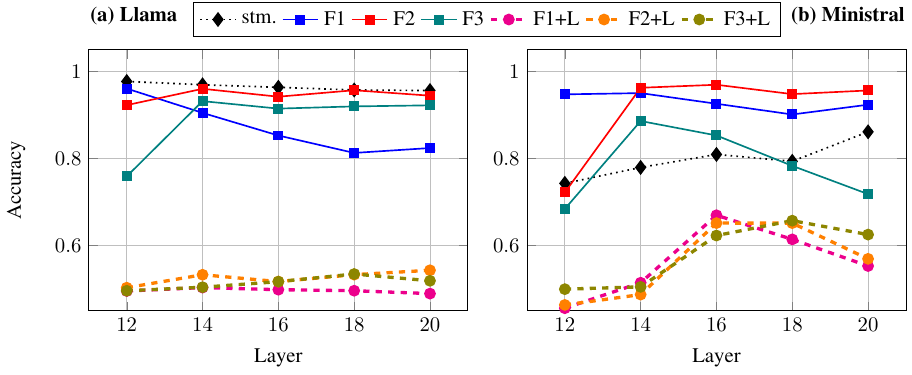}
    }}
    \caption{Generalization accuracy for the LRC probe trained on the \textbf{statements} (stm) dataset to other formats, for Llama-3-8b-instruct (a) and Ministral-8b-instruct-2410 (b).}\label{fig:Figure2}
\end{figure*}

\textbf{From statements to conversations:} We start this section with taking a look at how well the original dataset of true/false statements used in \cite{burger2024truth} generalizes to conversational formats, depicted in Fig. \ref{fig:Figure2}. We can clearly see that the truth direction learned from statements exhibits surprisingly good generalization to other conversational formats (F), but very poor generalization to longer conversational formats where the lie is buried slightly deeper in the prompt (F+L). It is important to emphasize that the results in Fig.~\ref{fig:Figure2} are not limited to logistic regression probe used here. For example, we observe similar poor generalization of the TTPD probe from~\cite{burger2024truth} to longer formats that do not end with a lie. Finally, note that Ministral has better performance on the conversational formats, even when trained on the statements, which we expect to be related to the instruct version.  

\begin{figure*}[htbp!]
    \centering
    \makebox[\textwidth][c]{%
    \resizebox{\textwidth}{!}{%
    \includegraphics[width=\textwidth]{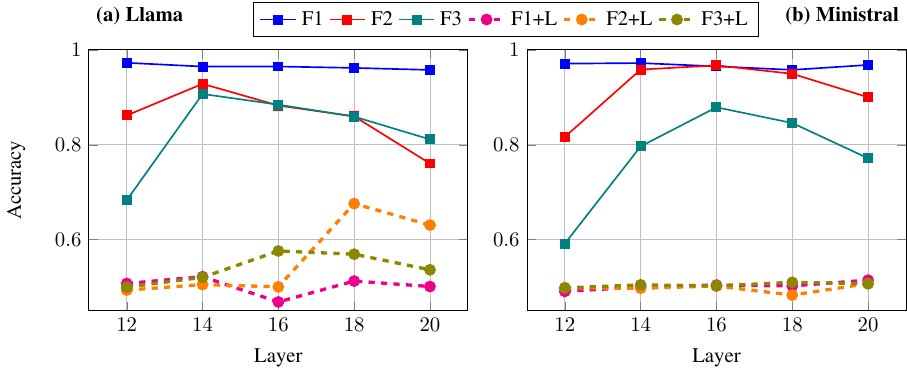}
    }}
    \caption{Generalization accuracy for the LRC probe trained on the \textbf{F1} dataset to other formats, for Llama-3-8b-instruct (a) and Ministral-8b-instruct-2410 (b).}\label{fig:Figure3}
    \label{fig:Figure3}
\end{figure*}

\textbf{Longer conversations remain problematic:} To eliminate the possible ambiguity of generalizing from statements to conversations, we rerun the same experiment, but now train on format F1 rather than the original dataset. In Fig.~\ref{fig:Figure3} we present the generalization accuracy across all layers from F1 to the other formats. Just as before, we observe poor generalization to any of the F+L formats. 

\begin{figure*}[htbp!]
    \centering
    \makebox[\textwidth][c]{%
    \resizebox{\textwidth}{!}{%
    \includegraphics[width=\textwidth]{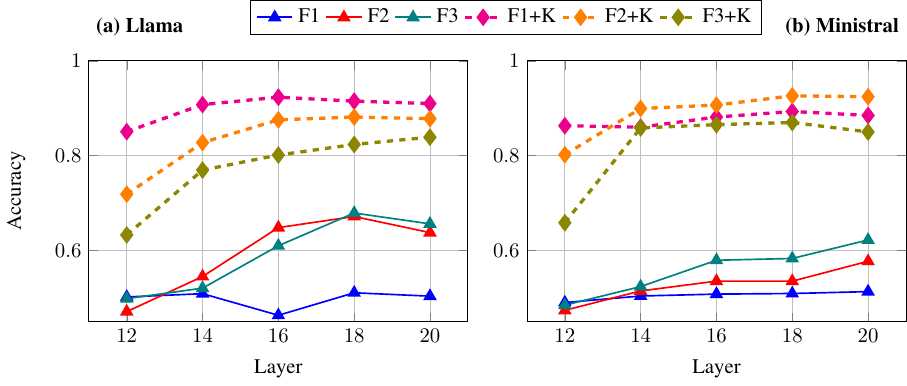}
    }}
    \caption{Generalization accuracy for the LRC probe trained on the formats shown in the legend, and tested on that same format but with the longer (+L) part included. For example, F1 scatters represent F1 $\rightarrow$ F1+L, and F2+K scatters represent F2+K $\rightarrow$ F2+L+K. Results shown for Llama-3-8b-instruct (a) and Ministral-8b-instruct-2410 (b).}\label{fig:Figure4}
\end{figure*}

\textbf{Adding a key phrase:} This leads us to make the hypothesis that adding some unrelated small talk after the lie weakens the attention of the last hidden state to the part of the prompt that contains a lie. To try and improve this we add a \textit{key phrase} as depicted on Fig.~\ref{fig:Figure1}, which provides an objective that asks the LLM to see if the conversation contains a lie. The motivation is that this would keep the attention of the model focused on the lie, even when it is buried deeper in the prompt. 

\begin{figure*}[htbp!]
    \centering
    \makebox[\textwidth][c]{%
    \resizebox{\textwidth}{!}{%
    \includegraphics[width=\textwidth]{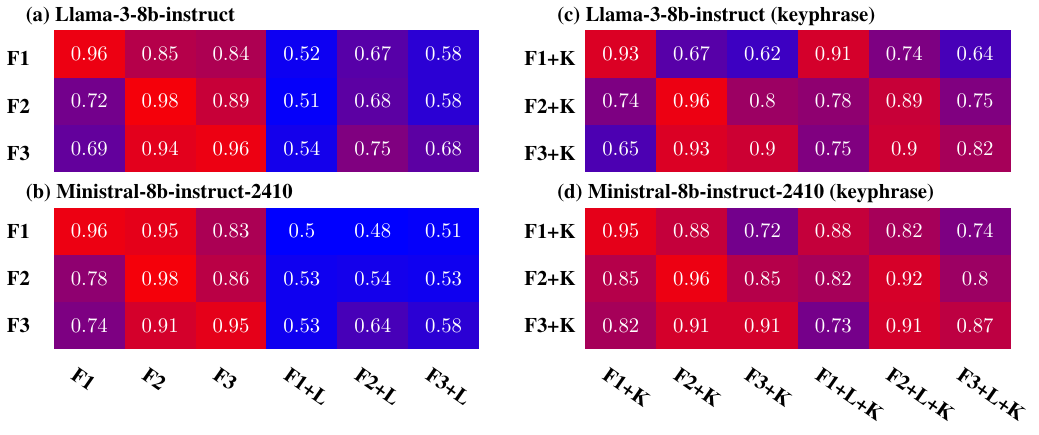}
    }}
    \caption{Cross-format generalization accuracy matrices from layer 18, for respectively no keyphrase (left column) and every format ending on a keyphrase (right column). The rows of the matrices represent the training sets, and the columns the test sets.}\label{fig:Figure5}
\end{figure*}

In Fig.~\ref{fig:Figure4}, we test this hypothesis by comparing the generalization from the short to long formats without (F $\rightarrow$ F+L) and with the key phrase (F+K $\rightarrow$ F+L+K). We can clearly see that in the presence of the key phrase, a major improvement is observed for every format.  
We can also take a closer look at the various cross-format generalization accuracies at a fixed layer in 
Fig.~\ref{fig:Figure5}. This confirms the previous result and shows that it also holds up towards \textit{any} F+L test set, even those different from the training format. On the other hand, this does not hold up for the generalization within the $F_i \rightarrow F_j$ formats, and even some smaller drops in accuracy can be observed. This suggests a trade-off, where these improvements come at the cost of a slight decrease in performance on the base formats. Nevertheless, when considered on average (for every row), using the key phrase still gives significant improvements. 

\begin{figure*}[htbp!]
    \centering
    \includegraphics[width=0.95\textwidth]{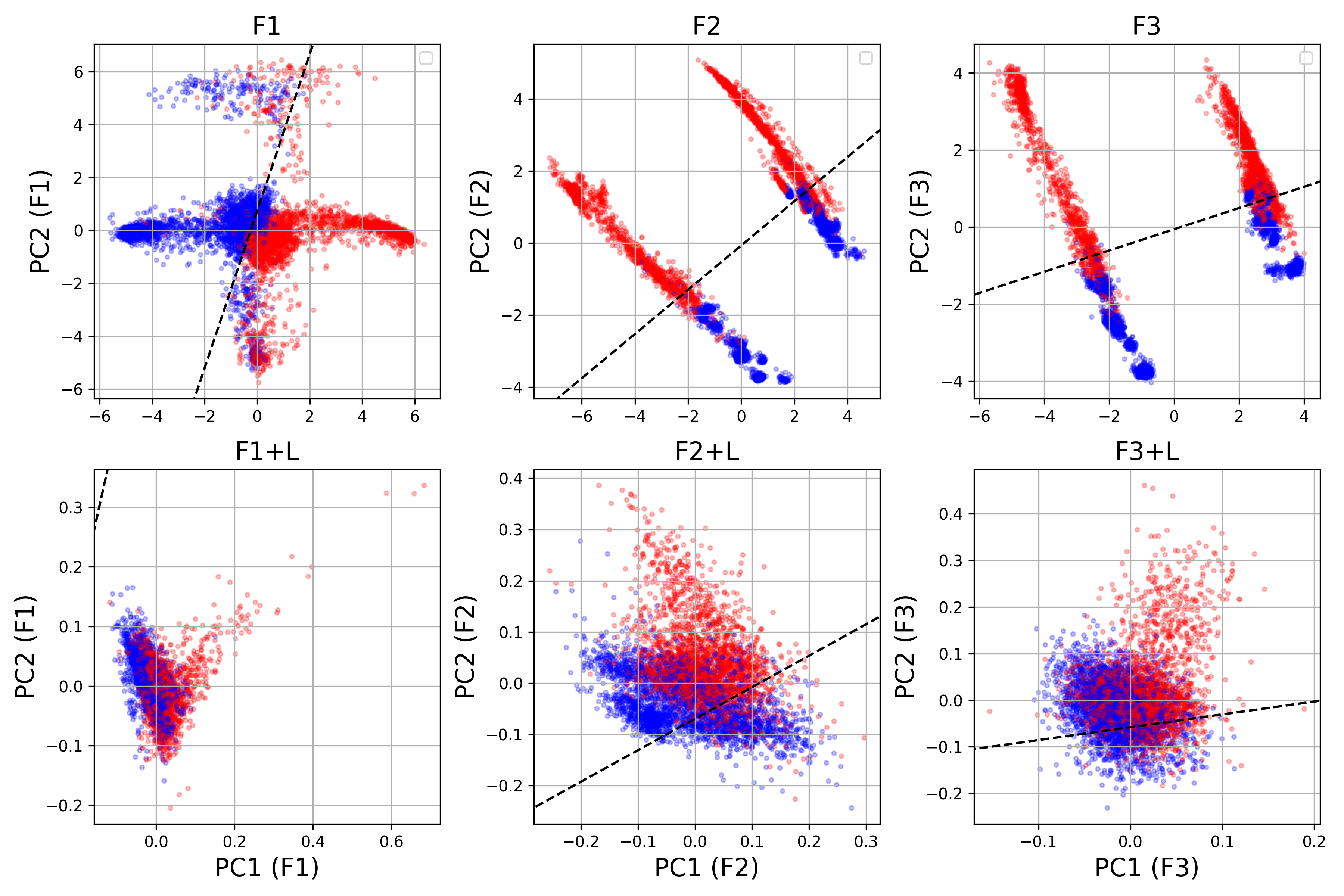}
    \caption{Visualization of the generalization F $\rightarrow$ F+L in layer 18 of \textit{Llama-3-8b-instruct}. For the centered activation data sets of every F format we determine the two main PCA components and project the activations on this plane (first row), and then also project the corresponding F+L centered activations on that same PCA plane (second row). The true/false labels are shown in blue/red, and the decision boundary of a probe trained on each of the F formats (first row) is projected on the second row as well. 
    }\label{fig:Figure6}
\end{figure*}

\begin{figure*}[htbp!]
    \centering
    \includegraphics[width=0.95\textwidth]{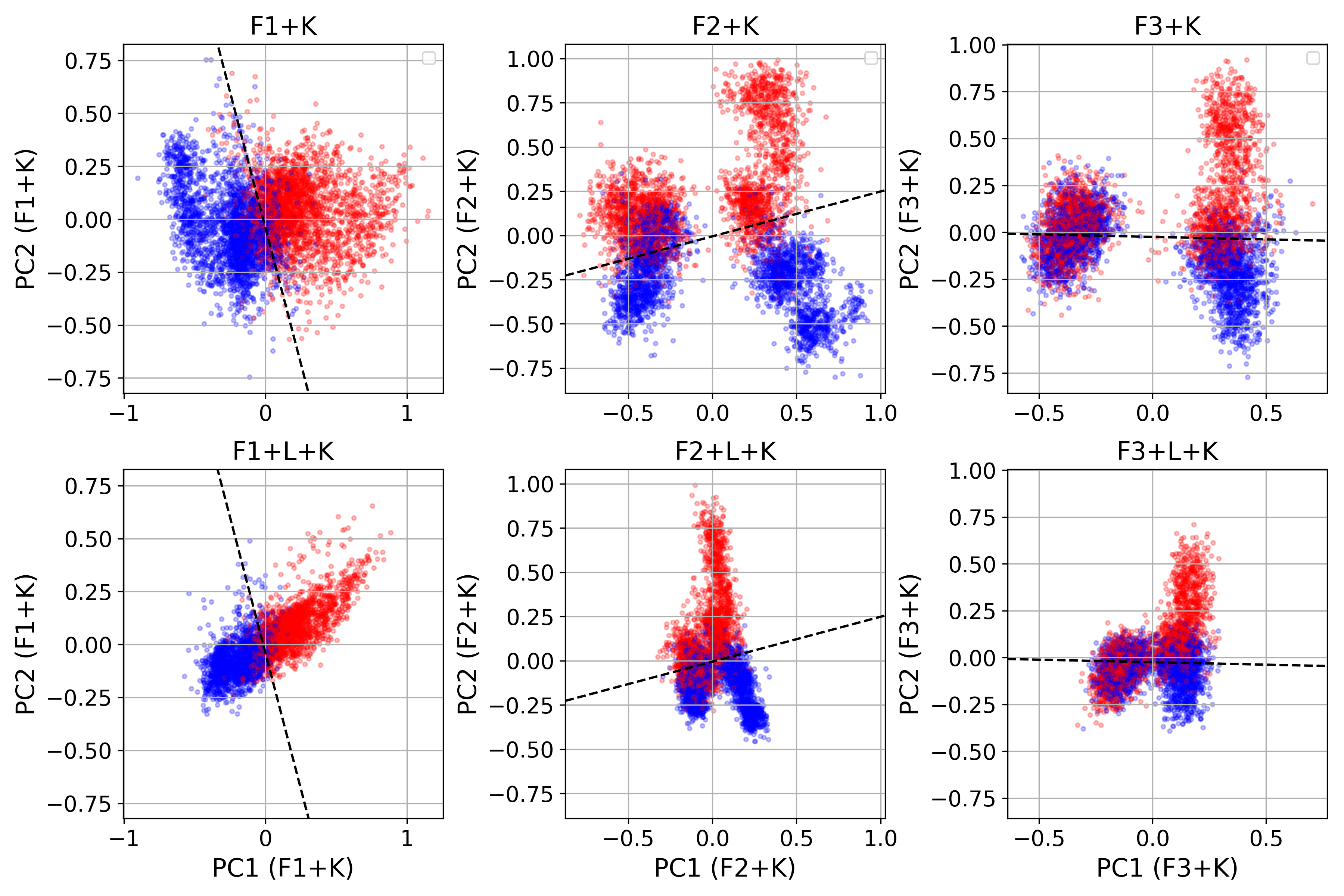}
    \caption{Similar visualization to Figure \ref{fig:Figure6}, but now for the generalization F+K $\rightarrow$ F+L+K with the key phrase added.}
    \label{fig:Figure7}
\end{figure*}

\textbf{Visualizing principal components:} To provide some visualizations to aid the previous discussion, we select a fixed layer for \textit{Llama-3-8b-instruct} and show several PCA plots in Fig.~\ref{fig:Figure6}. In particular, we determine the 2D PCA plane for every standard format F, and then project the activations of the longer formats F+L on the respective planes. We can clearly see that the projections of F+L look drastically different and that the decision boundary of the LRC probe has a poor generalization. Also note that we had to zoom in on the F+L data to make it more discernible on the figure. On Fig.~\ref{fig:Figure7}, we rerun the experiment but now look at the projections of F+L+K activations on the respective F+K PCA planes. Here, we see a stark contrast with the previous result and observe a far more similar data shape, size and structure when projected on the F+L PCA planes. The presence of the key phrase therefore seems to keep the structure of the activations far more robust when making the conversation longer, which would be a desired property for more reliable lie detectors. This provides an additional visual confirmation to the results in this work.

\begin{figure*}[htbp!]
    \centering
    \makebox[\textwidth][c]{%
    \resizebox{\textwidth}{!}{%
    \includegraphics[width=\textwidth]{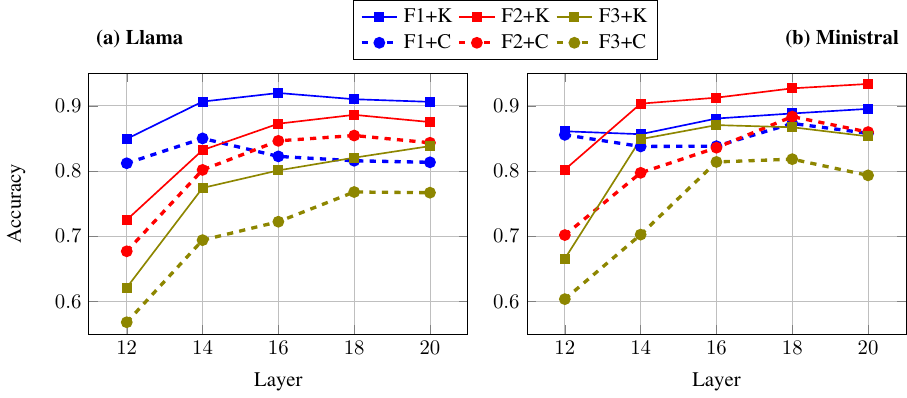}
    }}
    \caption{Generalization accuracy from short to long while using a key phrase (F+K$\rightarrow$F+L+K) on each of the three formats, comparing between the key phrase used the rest of this work (+K) and a control key phrase (+C).}\label{fig:Figure8}
\end{figure*}

\textbf{Comparing with a control key phrase:} We can also compare the effect of the key phrase that was used in this work (asking the LLM to classify the conversation for truth and lies) as provided in Fig.~\ref{fig:Figure1} with a control key phrase (asking the LLM to count the number of letters in the conversation). For the two key phrases we compare the generalization from the standard formats to the longer formats in Fig.~\ref{fig:Figure8}. As expected, we can see that the key phrase which asks about truths and lies consistently beats the unrelated control phrase. However, when we compare with Fig.~\ref{fig:Figure4} we see that the control key phrase still performs better than the absence of any key phrase. This suggests that before focusing the LLM attention back on specifically truth and lies, simply standardizing the prompts already yields noticeable improvements. 

\vspace{20pt}

\section{Conclusion and outlook}\label{sec:conclusion}

In this work we explored the generalization properties of LLM truth directions \cite{marks2024the} in the context of conversational prompts. Our study follows up on one of the questions raised in \cite{burger2024truth}, where some evidence was presented in favor of this generalization. Our main finding is that a critical generalization failure is observed when moving towards slightly longer prompts where the lie is not located at the very end. We also propose a remedy and show that adding a key phrase at the end of each input drastically improves this generalization. The results in this work highlight the difficulty of achieving broadly generalizable LLM lie detectors. Although we provide a solution to the generalization problem for this particular case, many more dimensions of generalizability remain to be explored. 

It is important to address the potential limitations of our work that would need to be explored in future research. We rely on two relatively small LLMs of 8 billion parameters, and it is not clear how our conclusions will hold up towards much larger models. As discussed in the introduction, we center all the activations, and hence the observations made in our work are valid for the learned truth \textit{direction} or \textit{plane} but ignore the bias. This means that the LRC probe cannot be directly used in a real-world scenario with only a single instance in the test set and one needs to either find a way to infer the appropriate centering or train a format-dependent bias. Finally, the datasets in our study contain relatively simple lies and more complex constructions should be explored in the future. We believe that one particularly interesting research direction would be to explore whether synthetic LLM-generated conversations by a larger model can help to overcome some of these challenges and train probes for smaller models.

\section{Acknowledgments}\label{sec:relwork}

We acknowledge the support of the “Flemish AI Research Program” (FAIR), and the Research Foundation Flanders (FWO), Grant G0G2721N. 

\bibliographystyle{unsrtnat}  
\bibliography{refs}

\end{document}